\documentclass{article}

\usepackage{arxiv}

\usepackage[utf8]{inputenc} % allow utf-8 input
\usepackage[T1]{fontenc}    % use 8-bit T1 fonts
\usepackage{hyperref}       % hyperlinks
\usepackage{url}            % simple URL typesetting
\usepackage{booktabs}       % professional-quality tables
\usepackage{amsfonts}       % blackboard math symbols
\usepackage{nicefrac}       % compact symbols for 1/2, etc.
\usepackage{microtype}      % microtypography
\usepackage{lipsum}		% Can be removed after putting your text content
\usepackage{graphicx}
\usepackage{doi}

\title{Changing Data Sources in the Age of\\Machine Learning\\for Official Statistics}

\date{June 6, 2023}	% Here you can change the date presented in the paper title
\author{ \href{https://orcid.org/0000-0003-0763-8114}{\includegraphics[scale=0.06]{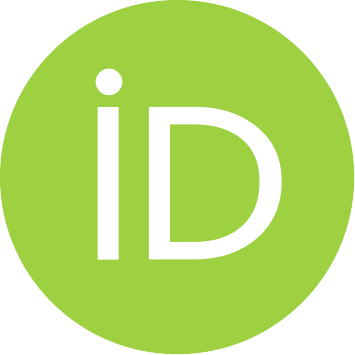}\hspace{1mm}Cedric De Boom} \\
	Statistics Flanders\\
	Belgium \\
	\texttt{cedric.deboom@vlaanderen.be} \\
	\And
	\href{https://orcid.org/0000-0002-9796-2016}{\includegraphics[scale=0.06]{orcid.pdf}\hspace{1mm}Michael Reusens} \\
	Statistics Flanders\\
	Belgium \\
	\texttt{michael.reusens@vlaanderen.be} 
}

\renewcommand{\headeright}{}
\renewcommand{\undertitle}{Presented at \textsc{UNECE} Machine Learning for Official Statistics Workshop 2023}
\renewcommand{\shorttitle}{}

\hypersetup{
pdftitle={Changing Data Sources in the Age of Machine Learning for Official Statistics},
pdfauthor={Cedric De Boom, Michael Reusens}
}

\begin{document}
\maketitle

\begin{abstract}
	Data science has become increasingly essential for the production of official statistics, as it enables the automated collection, processing, and analysis of large amounts of data. With such data science practices in place, it enables more timely, more insightful and more flexible reporting. However, the quality and integrity of data-science-driven statistics rely on the accuracy and reliability of the data sources and the machine learning techniques that support them. In particular, changes in data sources are inevitable to occur and pose significant risks that are crucial to address in the context of machine learning for official statistics.  

This paper gives an overview of the main risks, liabilities, and uncertainties associated with changing data sources in the context of machine learning for official statistics. We provide a checklist of the most prevalent origins and causes of changing data sources; not only on a technical level but also regarding ownership, ethics, regulation, and public perception. Next, we highlight the repercussions of changing data sources on statistical reporting. These include technical effects such as concept drift, bias, availability, validity, accuracy and completeness, but also the neutrality and potential discontinuation of the statistical offering. We offer a few important precautionary measures, such as enhancing robustness in both data sourcing and statistical techniques, and thorough monitoring. In doing so, machine learning-based official statistics can maintain integrity, reliability, consistency, and relevance in policy-making, decision-making, and public discourse.
\end{abstract}

% keywords can be removed
% \keywords{First keyword \and Second keyword \and More}

\section{Introduction}

The field of statistics has long played a critical role in informing policy decisions, driving innovation, and advancing scientific knowledge.
Traditional statistical methods such as surveys and censuses have provided valuable insights into a wide range of topics, from population demographics to economic trends and public opinion.
However, in recent years, the increasing availability of open and large data sources has opened up new opportunities for statistical analysis.
In particular, the rise of machine learning has transformed the field of statistics, enabling the analysis of massive datasets, the identification of complex patterns and relationships, non-linear forecasting, etc. \cite{russell_norvig_2009, hastie_friedman_tisbshirani_2017}.
Machine learning algorithms can be used to analyze data from a wide range of sources, providing insights that traditional survey methods may not capture.

The use of machine learning for official statistics has the potential to provide more timely, accurate and comprehensive insights into a wide range of societal topics \cite{unece2022}.
By leveraging the vast amounts of data that are generated by individuals and entities on a daily basis, statistical agencies can gain a more nuanced understanding of trends and patterns, and respond more quickly to emerging issues.

However, this shift towards machine learning also presents a number of challenges.
In particular, there are concerns about data quality, privacy, and security, as well as the need for appropriate technical skills and infrastructure \cite{Hassani2014, Puts2021}, as well as challenges related to explainability, accuracy, reproducibility, timeliness, and cost effectiveness \cite{Yung2022}.
As statistical agencies grapple with these challenges, it is essential to ensure that the benefits of machine learning are balanced against the risks and that the resulting insights are both accurate and representative.
In this paper, we explore the changing data sources in the age of machine learning for official statistics, as we believe that this pervasive issue largely remains underexposed, as we will explain in Section \ref{sec:external_data_sources}.
In that respect, we highlight some of the key considerations for statistical agencies looking to incorporate machine learning into their workflows in Section \ref{sec:changing_data_sources}, by zooming in on the causes and risks associated with using external data sources, the consequences on using such sources for statistical production, and, finally, a set of mitigations that should ideally be incorporated in any genuine deployment of machine learning for official statistics.

\section{Machine learning for official statistics}
The data abundance in governmental, corporate, social and personal contexts, both online and offline, becomes a tantalizing source and opportunity for the improvement and expansion of official statistics.
For example, to inquire about the overall satisfaction with life of its citizens, a nation could organize periodic surveys.
But when this nation has access to its citizens' social media posts, likes, reader's letters, media consumption, ticket sales, (online) shopping carts, etc.~it could use all of these data as a proxy to extract novel and innovative statistical insights \cite{Yung2021}.
Typically, the end result is either a new statistic, a statistic that complements an existing one, or an \textit{ersatz} statistic that aims to replace one or more existing statistics.
We will briefly zoom in on the different components of which such a novel statistic is generally comprised.

\subsection{Machine learning}
To derive novel insights and innovative statistics from data, data scientists and statistical researchers often use a wide variety of powerful tools that are in the realm of machine learning\footnote{Although originally (and technically) they imply different methods and techniques, the terms machine learning, data science, artificial intelligence, deep learning...~are nowadays considered interchangeable. In this paper we consistently use the term machine learning to denote the scientific discipline concerned with learning the most optimal model parameters based on data. It is a subdiscipline of artificial intelligence, while deep learning is a subdiscipline of machine learning. Data science encompasses both machine learning as well as data preparation, analytics and visualization.}.
In this paper we will not go into too much detail about machine learning.
However, the use of data sources together with machine learning models can cause unwanted effects regarding statistical production, see further in Section \ref{sec:changing_data_sources}.
So, we deem it useful to provide a high-level overview of the typical process that involves machine learning for official statistics.

Machine learning is a subdiscipline of artificial intelligence that enables machines to learn from data and improve their performance over time without being explicitly programmed.
This approach involves building algorithms that automatically learn from data to identify patterns, relationships, and structures that may be difficult or impossible for humans to discern.
The typical process of designing a machine learning algorithm consists of two main phases: training and inference. 
During the training phase, the parameters of a machine learning model are tuned to solve a specific task.
For this, a wide variety of data sources can be used, or even outputs from existing or pre-trained machine learning models.
After the model is trained, its parameters are kept fixed so that the model can be used to predict outcomes or identify patterns in new or previously unseen data.
This process is called inference.
It is important to keep in mind the distinction between training and inference.
After training, the model remains unchanged, and it remains unchanged until it is retrained again.
Later, in Section \ref{sec:changing_data_sources}, we will focus on the disparities that this can cause w.r.t.~the inference phase and, in consequence, official statistics production.

Supervised learning is one of the most common types of machine learning used for official statistics.
This method involves training a model on a labeled dataset, where each data point has a known outcome or target variable. 
The model learns to associate features in the data with the target variable, enabling it to make predictions on new data with similar features.
In the context of official statistics, supervised learning can for example be used to predict the happiness of an individual based on their Twitter profile \cite{reusens2022}.
Unsupervised learning, on the other hand, is used when the target variable is unknown or the goal is to identify patterns or relationships within the data.
In this approach, the machine learning model learns to recognize similarities and differences among input data without explicit guidance from labeled data.
In the context of official statistics, unsupervised learning can for example be used to identify citizens, companies or events that are similar to each other on one or more aspects that could be hidden from plain sight \cite{crijns2023}.

Machine learning can be used to complement or even replace official statistics, and its ability to nowcast and forecast is an extremely valuable addition.
Modern machine learning models, tools and hardware can analyze vast amounts of data in real-time or near-real-time, providing more up-to-date and precise estimates of e.g.~economic and social trends.
By incorporating machine learning into official statistical production, one can benefit from the strengths of both approaches and make more informed decisions based on the most current and accurate data \cite{Erman2022}.

\subsection{External data sources}
\label{sec:external_data_sources}
Let's focus on the data sources that will power such machine learning models.
Their nature, size, structure, frequency...~can be vastly different, they must typically be gathered `in the wild' and should often be combined with each other to extract meaningful insights.
Compared to more traditional data sources for official statistics, they may present unique and appealing characteristics such as:\\[-0.2cm]

\noindent\textbf{Broad-spectrum} -- Covers a wide variety of topics.\\[-0.2cm]

\noindent\textbf{Diversity} -- A large variety of sources to cover different perspectives.\\[-0.2cm]

\noindent\textbf{Availability} -- Lots of data is freely and easily accessible.\\[-0.2cm]

\noindent\textbf{Size} -- Some datasets can be enormous, sometimes even complete.\\[-0.2cm]

\noindent\textbf{Structure} -- Not only tabular data, but also images, video, text, audio, etc.\\[-0.2cm]

\noindent\textbf{Timeliness} -- (Near) up-to-date and real-time information.\\[-0.2cm]

\noindent\textbf{Frequency} -- Raw data on various, even very fine-grained time scales.\\[-0.2cm]

\noindent\textbf{Granularity} -- Raw data on various, even fine-grained levels of detail.\\[-0.2cm]

\noindent\textbf{Coverage} -- Various locations and regions can be filtered and covered.\\[-0.2cm]

On the other hand, before all this data is ready to be exerted for machine learning and official statistical production, a few challenges need to be overcome, such as:\\[-0.2cm]

\noindent\textbf{Data quality} -- Data may contain errors, biases, or missing values that need to be addressed to ensure accuracy and reliability.\\[-0.2cm]

\noindent\textbf{Data interpretation} -- Understanding the context and meaning of data can be difficult, especially when dealing with unstructured data such as text or images.\\[-0.2cm]

\noindent\textbf{Data integration} -- Combining data from different sources with varying structures and formats can be challenging and time-consuming.\\[-0.2cm]

\noindent\textbf{Selection bias} -- Proper randomization or compiling representative population samples can be challenging, and it greatly depends on the underlying data origins.\\[-0.2cm]

\noindent\textbf{Operationalization bias} -- Reproducibility can be difficult as it depends on many implicit, hidden, and/or production-specific design choices \cite{Haucke2021, Reddy2022}.\\[-0.2cm]

\noindent\textbf{Computational resources} -- Processing and analyzing large amounts of data may require significant computational resources.\\[-0.2cm]

\noindent\textbf{Privacy and security} -- Sensitive data may need to be protected and anonymized to ensure privacy and security.\\[-0.2cm]

\noindent\textbf{Data ethics} -- Data collection and use should adhere to ethical principles.\\[-0.2cm]

\noindent\textbf{Fairness and justness} -- The end solution should ideally be as neutral as possible and should not discriminate \cite{Kuppler2022}.\\[-0.2cm]

\noindent\textbf{Cost} -- All of the above requires resources, budgets and a talented workforce. In addition, the data source itself might need to be purchased. In 2016, McKinsey reported that many companies have started to specialize in acquiring and selling data \cite{mckinsey2016}.

With the right tools, workforce, technological advances, mindset, and legislative support, these challenges can and should be manageable.
The most challenging piece of the puzzle, however -- and one that is more than often ignored -- is the \textit{lack of control} you can exert over the data sources that are externally gathered.
As a national statistics agency, traditionally, survey data and administrative records that power official statistics are completely under your own control.
But once you start exploiting external data sources to power novel, innovative, complimentary or ersatz statistics, this lack of control of your data should never be ignored, and if possible, should be front and center on your agenda early on in the process.

As the popular saying goes: ``With great power comes great responsibility'' (from Spider-Man, 2002).
Having control and power over your data is essential to fulfilling your responsibilities as a statistics agency.
However, in the world of data, the opposite is often true: with great amounts of external data comes great powerlessness.
Therefore, it is crucial to prioritize the issue of data control when incorporating external sources into official statistics.
Taking the time to establish proper protocols and procedures for external data management can prevent a multitude of issues down the line and ensure that the data you rely on remain accurate and trustworthy.

This paper delves into the pervasive problem of powerlessness and lack of control, unraveling the multifaceted aspects, risks, and pitfalls that arise from utilizing external data sources for machine learning in official statistics.
We will explore the concepts of `change' and `consequence' in their most expansive interpretations to comprehensively tackle this question.

\section{The challenge of changing data sources}
\label{sec:changing_data_sources}

Relying heavily on external data sources for machine learning in official statistics comes with significant risks.
Such a dependence can leave statistical agencies vulnerable since they have limited control over these sources.
This situation is similar to how our global economy, mobility, and prosperity were once highly dependent on the availability of oil.
Since the prices and availability of these precious resources are often beyond our control, countries can do nothing but endure price fluctuations and shortages.
Clive Humby proclaimed in 2006 that ``data is the new oil'', given its powerful intrinsic value.
However, his statement keeps holding true in terms of vulnerability, powerlessness, and lack of control over external providers.

In the following paragraphs, we will delve into the various types and causes of data changes.
We will then discuss the ramifications of changing data sources for machine learning in official statistics.
Finally, we will provide a list of best practices and tips, although it is important to remember that there is no free lunch: whenever we incorporate external data, we expose ourselves to the risk of future changes in these data sources.

\subsection{Types and Causes of Changing Data Sources}
\label{sec:types_and_causes}
\subsubsection{Data types and schemas}
A change in data types or schemas refers to modifications made to the data formats or the structure in which the data are stored and offered.
These types of changes may arise due to a need to accommodate future use cases or business requirements, to eliminate technical debt, or to improve data storage and retrieval efficiency.
Even the most innocent changes -- e.g.~integers becoming floats, data columns that are added or removed... -- can break entire pipelines.
In the most fortunate of cases the runtime environment will throw errors that reveal the cause of these data changes.
In other cases, however, the data changes remain undetected and secretly wreak havoc in the pipeline.
If the pipeline contains machine learning components, data type changes can e.g.~induce feature mismatches -- discrepancies between the feature distributions at train and inference time -- that lead to unreliable predictions.

It is important to be vigilant about changes in data types or schemas, as even seemingly minor adjustments can have significant impacts further down the data pipeline.
To mitigate these risks, it is advisable to stay informed about data change announcements from providers and implement robust data checks during data ingestion, ranging from simple data (type) validation to full-blown automated feature analysis, outlier detection, etc.
Additionally, the deployment of effective monitoring systems can help catch machine learning failures quickly and prevent potentially costly errors.

\subsubsection{Sharing and collection technology}
Data can be shared and collected using many different technologies, such as APIs, queues, network drives, external drives, e-mail...~but also web scraping, online analytics tools, sensor networks...
Changes in these technologies inevitably occur from time to time.
For example, API endpoints often need to be updated to improve functionality and performance.
Changes may be made to the API's data structures or methods, to provide more efficient or comprehensive data access.
In addition, changes may be made to the API to address security vulnerabilities or to ensure compliance with new regulations or standards.
Furthermore, changes in business requirements or strategy may also lead to changes in API endpoints.
For instance, a company may introduce new products or services, modify their existing offerings, or change their pricing.

A recent, telling example is the Twitter API.
In 2021, Twitter launched version 2 of its popular API that introduced many changes in endpoints, data fields, pricing...~compared to version 1.1.
Twitter encouraged developers to migrate to this new API offering, but for many use cases such a migration would introduce breaking changes that, in their turn, would impact entire data processing pipelines, statistics production, etc.
For the time being, Twitter offered both version 1.1 and version 2 of their API in parallel, which caused many to bury their heads in the sand.
The situation got even worse when Elon Musk acquired Twitter in 2022 and decided to suspend all existing API offerings.
Instead, in 2023 a new enterprise tier was introduced that put a price of more than 40 thousand USD per month on any reasonably effective use of the API.
This caused great dissatisfaction in the development and research community and many initiatives were abandoned.

\subsubsection{Concept drift}
Concept drift is related to changes in the data distribution between train and test time, which can have multiple causes \cite{Hu2020, Bayram2022}.
Changes in business logic can induce information shifts, for example, when categorical variables are expanded with additional categories or when the meaning of certain data fields is altered.
A particular pervasive issue is the calculation of derived data fields, especially when those calculations are not transparent or proprietary.
In the age of machine learning, you should always assume that derived data fields can be the result of a model prediction; when this model is updated without your knowing, the derived data fields will have a (slightly) different data distribution, which will cause issues in dependent machine learning models.
But even when data fields are not the result of a model's prediction, it is important to periodically reevaluate and retrain models, since many sociological and economic processes are naturally prone to concept drift themselves.

\subsubsection{Frequency and interruptions}
A change in data frequency refers to modifications made to the rate at which data is collected or updated, which can happen deliberately or randomly.
Deliberate changes may arise due to shifts in business requirements or technological choices.
Random shifts are most often attributed to noisy factors such as network issues, component failures, downtime...~or (human) errors.
Such changes in frequency can impact machine learning components dramatically.
E.g.~when periodical data is sampled every minute instead of every second, the data distribution changes on which the model was trained.
To mitigate these risks, data pipelines should be designed to monitor changes in incoming data frequency.

\subsubsection{Ownership and discontinuation}
Worse than interruptions is downright discontinuation of the data source, which has immediate consequences on the future existence of the statistic.
Also, a change of ownership of the data source -- e.g.~when acquired by another company -- is not a fictional scenario, and it can trigger any of the risks that are discussed in this section.
Building redundancy by diversifying data sources is a useful mitigating strategy to avoid single points of failures.

\subsubsection{Legal properties}
Legal changes refer to modifications made to the legal landscape that governs the collection, storage, and use of data.
This type of change may arise due to new privacy laws, contractual obligations, or changes in the cost of data access or storage.
One cause of this type of change is the adoption of new regulations, such as GDPR, which require companies to comply with stricter rules for collecting and processing data.
Additionally, changes in the cost of data access or storage may require companies to modify their data sources or methods to reduce costs, which can have contractual consequences.
If possible, negotiate airtight SLAs with the data provider and make sure to attribute enough attention to future data changes.

\subsubsection{Ethics and public perception}
Ethical considerations and public perceptions can affect data collection methods and sources.
If certain data sources or variables are considered controversial or intrusive, there may be a shift towards alternative sources, which may require a refresh of the used machine learning models.
It can also impact the way machine learning models are designed and trained.
If certain variables or factors are considered discriminatory or unethical, there may be a push towards eliminating or adjusting them to reduce algorithmic bias.
Changes in ethics or public perceptions can also result in greater accountability and the need for transparency.
Stakeholders may demand more openness and clarity around the use of algorithms, data sources, and decision-making processes.
This can lead to greater scrutiny and oversight of machine learning models, which may impact their performance if not adequately addressed, especially when black-box models need to be replaced by more interpretative variants \cite{Gosiewska2021, Rudin2019, London2019}.
Finally, public trust can be significantly affected.
If stakeholders perceive that machine learning is being used inappropriately or unethically, they may lose faith in the integrity and reliability of official statistics. This can have significant consequences for public policy and decision making.

\subsection{Consequences of Changing Data Sources}
When data sources change, there will be consequences for official statistics production, especially if there are machine learning components involved.
We will broadly but briefly cover a variety of areas that can be impacted, some of which have already been mentioned above.\\[-0.2cm]

\noindent\textbf{Concept drift} -- Concept drift means that the underlying patterns and relationships in the data may change over time, which can lead to model deterioration or loss of accuracy.
This issue can be particularly relevant when dealing with long-term trends, as changes in societal norms, technology, or other external factors can influence data over time \cite{Bayram2022}.\\[-0.2cm]

\noindent\textbf{Model staleness} -- When a model becomes outdated, it no longer reflects current trends or patterns in the data.
This can occur if the machine learning model is not updated frequently enough to keep pace with changing data sources.
As a result, the model may not perform as well as it once did, leading to less accurate official statistics.\\[-0.2cm]

\noindent\textbf{Bias and neutrality} -- Changing data sources can also introduce bias or incorrect data, which can impact the neutrality of the statistics produced, or which can lead to the phenomenon ``garbage in, garbage out'' \cite{Vidgen2020,Geiger2021}.
Since it is essential that official statistics remain neutral and objective, this will negatively impact the accuracy and validity of these statistics.\\[-0.2cm]

\noindent\textbf{Availability} -- If data become unavailable (for a certain period in time or indefinitely) or are limited in scope, this may impact the ability to produce accurate and timely official statistics.\\[-0.2cm]

\noindent\textbf{Integration} -- A change in data sources can cause a domino effect when multiple statistics or models rely on this data source.
Especially be mindful when the output of machine learning models is used as input for other machine learning models, either directly or indirectly as part of a larger data pipeline.
Since the predictions of a machine learning model can become unreliable when the input data change, this prediction shift itself is a changing data source for other models.\\[-0.2cm]

\noindent\textbf{Extra labor} -- The risk of changing data sources requires additional resources and labor to mitigate the effects of such changes, monitor the occurrence of changes, and ensure that the new data are properly integrated into existing machine learning models.
This has tremendous impacts on the costs and timeline of the produced statistics, and it may also require a significant team expansion.\\[-0.2cm]

\noindent\textbf{Breaking changes or discontinuation} -- In some cases, changing data sources may cause the impossibility of producing a statistic any further.
If this is the case, it may be necessary to stop offering the statistic altogether or find alternative data sources that can produce accurate and reliable official statistics.
When alternative data sources are found, there will almost always be a mismatch with the original data source that has an impact on the resulting statistic, resulting in a breaking change.
In that case, it is important to overcome the mismatches as best as possible -- e.g.~in terms of statistical properties -- and, certainly, to be transparent about the breaking change, e.g.~by indicating on a graph when exactly the data source was changed.\\[-0.2cm]

\noindent\textbf{Quality metrics} -- Finally, changing data sources imply changes in timeliness, validity, accuracy, completeness, consistency and other quality metrics w.r.t.~the produced official statistics \cite{Yung2022}.
Ensuring that resulting statistic continues to meet these quality metrics remains critical.

\subsection{Mitigating Changing Data Sources}
As has been illustrated above, changes in data sources can significantly impact the performance of a machine learning model.
The effects can be diverse, ranging from introducing biases in the data to producing incorrect results.
This can have serious implications, especially when the model is being used for official statistics, where accuracy and reliability are of paramount importance.
Therefore, it is essential to take measures to prevent and mitigate such changes.
This is not an easy task, as the consequences can be diverse, and the required efforts to mitigate them are often time-consuming and not straightforward.
We do not claim to have definite answers.
However, we will propose several recommendations and best practices, including performing a risk analysis, monitoring, diversifying data sources, building technical robustness, using data normalization techniques, and incorporating data validation processes.\\[-0.2cm]

\noindent\textbf{Risk analysis} -- Performing a risk analysis before incorporating a new data source is an essential step in mitigating the impact of changes in data sources.
This analysis involves identifying the potential risks associated with the data source, which we have covered in Section \ref{sec:types_and_causes}.
The analysis should be comprehensive, considering both technical and non-technical aspects of the data source, and should ideally include potential solutions for the identified risks.
This will often force you to face the hard truth and will lead you to decide that the candidate data sources are not adequate or reliable enough.
Trade-offs will nevertheless need to be considered, depending on the use case at hand.\\[-0.2cm]

\noindent\textbf{Monitoring} -- Monitoring everything that is relevant is another crucial step in mitigating the impact of changing data sources.
It involves tracking various aspects of the data sources, the machine learning models, and their outputs to detect and respond to changes promptly.
Draft a list of variables and quantities that must be continuously tracked to ensure that the models remain reliable and accurate over time.
For this, inspiration can be drawn from the discussed topics in Section \ref{sec:types_and_causes}, but it will vary from use case to use case, as well as the nature of the models that have been used.\\[-0.2cm]

Supervised models, for example, can be tested against a reference test set or a historical reference model; if the accuracy, precision or recall starts to deviate significantly from this reference set, it should be flagged.
On the other hand, monitoring the performance of unsupervised models can be more challenging, because there is no clear performance measure that can be directly computed.
One approach is to monitor the model's ability to detect patterns and clusters in the data.
It is possible to use a reference test set or reference model for this, but the informative metrics -- e.g.~cluster similarity, homogeneity, separation... -- are more abstract and somewhat harder to interpret.
Another approach is to visualize projections of certain interesting data points in the learned latent spaces or preferably a reduction thereof, which greatly benefits interpretability but makes it harder to convert it into hard numbers.
As a suggestion, a good balance between interpretability and hard performance metrics is found when clusters are tested against pre-existing domain knowledge, e.g.~by listing similar data points for given queries.
Simply monitoring whether expected similarities emerge or not can provide powerful signals about model and data performance.
Another effective approach is to create proxy supervised tasks that rely on the output of the unsupervised model.
Monitoring the model's performance on such proxy tasks can provide insights into the quality and usefulness of the unsupervised model's output.

\noindent\textbf{Diversification} -- Diversifying data sources is another important measure, but is easier said than done.
One challenge of using multiple data sources is the potential for conflicts or inconsistencies between the sources.
Different data sources may have different formats, schemas, and levels of quality, which can create discrepancies and inconsistencies that must be resolved before the data can be used in the model.
Therefore, data normalization is key.
Additionally, integrating multiple data sources can be a complex and time-consuming process.
It can also create additional computational overhead, which may impact the model's scalability and portability.
Finally, finding relevant and reliable data sources can be a challenging task, particularly for specialized or niche domains.
It may require extensive research and communication with data providers to retrieve relevant data.
Again, this story is about economical, technical and practical trade-offs, and is of course highly use-case-dependent.\\[-0.2cm]

\noindent\textbf{Technical robustness} -- Building technical robustness is paramount and requires significant engineering efforts.
Building an automated, data-driven statistic that is resistant to changing data sources such as errors, outliers, outages, time-dependent variability, etc.~ensures consistency in the statistical offering.
Using data normalization techniques and incorporating data validation into the pipeline are essential measures, but robust technical implementations also require thorough unit and integration testing, failover and deduplication, scalability solutions, security measures, etc.
Of course, this is an entire field of study on its own.\\[-0.2cm]

\noindent\textbf{Legal robustness} -- Finally, we believe that agreeing on clear legal guidelines is the best mitigation strategy to counter the risk of changing data sources, for example, by closing formal data sharing agreements or SLAa with data providers.
Such agreements should specify the terms and conditions under which the data can be shared, as well as the legal responsibilities of each party.
In particular, the agreements should specify the legal consequences of non-compliance.

\section{Conclusion}
In this paper we have investigated the risks and consequences of changing data sources when using machine learning for official statistics.
The list is long and covers many different aspects, ranging from statistical issues and model inconsistencies to technical problems and ethical considerations.
We have also looked at a few potential mitigation strategies.
However, we admit that these strategies do not provide all the adequate answers and might leave the reader unsatisfied or, worse still, beguiled, as the solutions require many additional resources and efforts.
As we have stressed a couple of times in this paper, this is a story of trade-offs.
Depending on the use case at hand, some trade-offs might be easier to handle than other ones.
However, in the context of official statistics, our advice is to not tread lightly on these matters and to minimize the risk of losing control over your data sources as much as possible.
This takes time, effort and careful planning with a horizon of multiple years.
To end on a positive note, despite the challenges associated with changing data sources, machine learning offers many opportunities for official statistics.
By being aware of the risks and taking necessary precautions, statistical agencies can leverage these opportunities while maintaining the integrity and reliability of their data-driven products.
We hope that our checklist of risks and mitigation strategies provides a useful starting point for statistical agencies and practitioners to ensure the robustness of their machine learning-based statistical reporting.

\bibliographystyle{unsrt}
% \bibliography{template.bib}  %%% Uncomment this line and comment out the ``thebibliography'' section below to use the external .bib file (using bibtex) .

%%% Uncomment this section and comment out the \bibliography{references} line above to use inline references.

\end{document}